\documentclass[10pt,twocolumn,letterpaper]{article}

\usepackage{cvpr}
\usepackage{times}
\usepackage{epsfig}
\usepackage{graphicx}
\usepackage{amsmath}
\usepackage{amssymb}
\usepackage{bm}
\usepackage{overpic}
\usepackage{multirow}
\usepackage{booktabs}       
\usepackage{array}
\newcommand{\PreserveBackslash}[1]{\let\temp=\\#1\let\\=\temp}
\newcolumntype{C}[1]{>{\PreserveBackslash\centering}p{#1}}
\newcolumntype{R}[1]{>{\PreserveBackslash\raggedleft}p{#1}}
\newcolumntype{L}[1]{>{\PreserveBackslash\raggedright}p{#1}}

\usepackage[pagebackref=true,breaklinks=true,letterpaper=true,colorlinks,citecolor=blue,linkcolor=blue,bookmarks=false]{hyperref}

\cvprfinalcopy 

\def\cvprPaperID{1251} 

\def\ie{i.e.,~} 
\def\blu#1{\textbf{\color{blue} #1}} 
\def\red#1{\textbf{\color{red}\underline{#1}}} 

\ifcvprfinal\fi
\begin{document}

\title{PiCANet: Learning Pixel-wise Contextual Attention for Saliency Detection}

\author{
	Nian Liu$^{1}$
	\hspace{25pt}
	Junwei Han$^{1}$\thanks{Corresponding author}
	\hspace{25pt}
	Ming-Hsuan Yang$^{2,3}$
	\\
	$^1$Northwestern Polytechincal University
	\hspace{8pt}
	$^2$University of California, Merced
	\hspace{8pt}
	$^3$Google Cloud
	\\
	{\tt\small 
    \{liunian228, junweihan2010\}@gmail.com
    \hspace{50pt}
    mhyang@ucmerced.edu
}
}

\maketitle
\thispagestyle{empty}
\pagestyle{empty}

\begin{abstract}
   Contexts play an important role in the saliency detection task. However, given a context region, not all contextual information is helpful for the final task. In this paper, we propose a novel pixel-wise contextual attention network, \ie the PiCANet, to learn to selectively attend to informative context locations for each pixel. Specifically, for each pixel, it can generate an attention map in which each attention weight corresponds to the contextual relevance at each context location.
 An attended contextual feature can then be constructed by selectively aggregating the contextual information.
   We formulate the proposed PiCANet in both global and local forms to attend to global and local contexts, respectively.
 Both models are fully differentiable and can be embedded into CNNs for joint training.
 We also incorporate the proposed models with the U-Net architecture to detect salient objects.
 Extensive experiments show that the proposed PiCANets can consistently improve saliency detection performance.
 The global and local PiCANets facilitate learning global contrast and homogeneousness, respectively.
 As a result, our saliency model can detect salient objects more accurately and uniformly, thus performing favorably against the state-of-the-art methods.
\end{abstract}
\vspace{-4mm}

\section{Introduction}

Saliency detection aims at modeling human visual attention mechanism to detect distinct regions or objects, on which people likely focus their eyes in visual scenes.
Contextual information plays an essential role in this visual task. As one of the earliest pioneering computational saliency models, Itti \etal \cite{itti1998model} calculate the feature difference between each pixel and its surrounding regions as the contrast to infer saliency.
 Numerous methods have been subsequently developed \cite{han2011bottom,cheng2015global,klein2011center} that utilize local or global contexts as the reference to evaluate the contrast of each image location (i.e., local or global contrast).
 These models aggregate visual information at all the locations of the referred context region into a contextual feature to infer contrast.

Recently, convolutional neural networks (CNNs) have been introduced into saliency detection to learn effective contextual representation.
Specifically, several methods \cite{li2015mdf,liu2015mrcnn,zhao2015mcdl} first directly use CNNs to extract features from multiple image regions with varying contexts and subsequently combine these contextual features to infer saliency. Some other models \cite{kuen2016recurrent,li2016dcl,liu2016dhsnet,liu2016dsclrcn,wang2016rfcn,hou2017dss,luo2017nldf,Zhang2017amulet,Zhang2017ucf,Wang2017srm} adopt fully convolutional networks (FCNs) \cite{long2015fcn} for feature representation at each image location and generate saliency map in a convolutional way.
In these models, the first school extracts contextual features from each input image region, while the second one extracts features at each image location from its corresponding receptive field.
\begin{figure}[!t]
  \graphicspath{{Figures/figure1/}}
  \centering
  \includegraphics[width=0.8\linewidth]{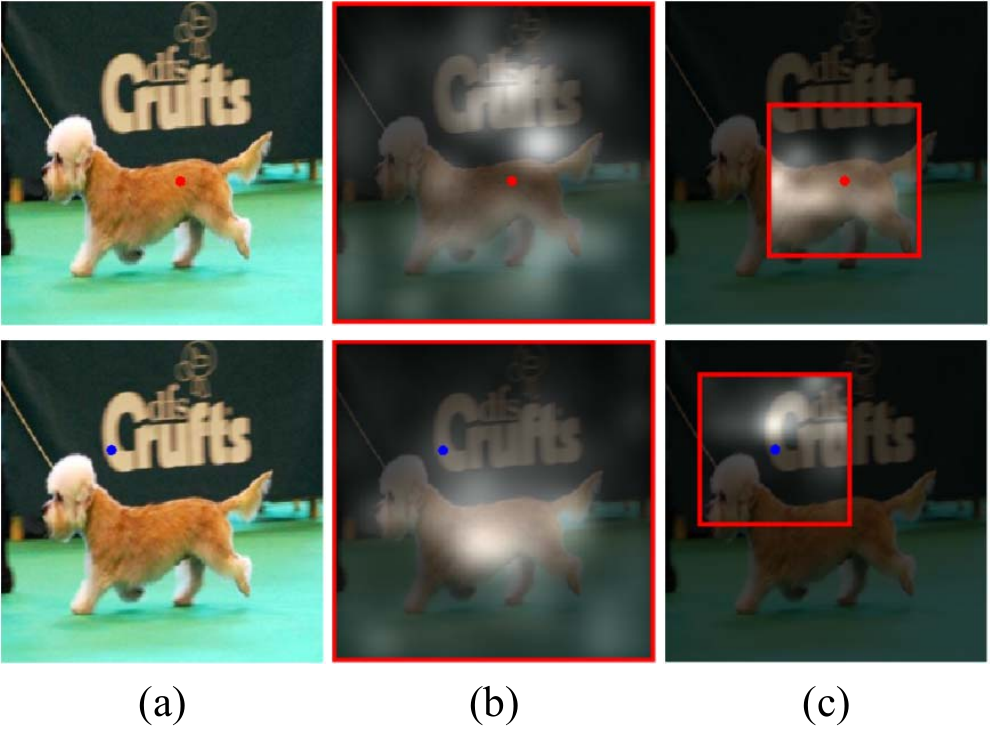}
  \caption{Example of learned global and local pixel-wise contextual attention maps. (a) shows the original image and two example pixels, \ie the red dot on the foreground dog and the blue dot on the background. (b) and (c) show the learned global and local contextual attention maps for the two pixels, respectively. The brightness of each location indicates the magnitude of its attention weight. The red boxes indicate the referred context regions.}
  \label{figure1}
  \vspace{-0.5cm}
\end{figure}

However, all the existing models utilize context regions holistically to construct contextual features, in which the information at every contextual location is integrated.
Intuitively, for a specific image pixel, not all of its contextual information contribute to its final decision.
Some related regions are usually more useful, while other noisy responses should be discarded.
For example, for the red dot pixel in the first row in Figure~\ref{figure1},
we need to compare it with the background to infer its global contrast.
If we want to check whether it belongs to the foreground dog for uniformly highlighting the whole dog, we need to refer to other parts of the dog.
While for the blue dot pixel in the second row, we need to refer to the foreground dog and other parts of the background, respectively.
Thus, if we can identify relevant context regions and construct informative contextual feature for each pixel, better decisions can be made.
Nevertheless, this important issue has not been addressed by existing methods.

To address the problem discussed above, in this paper we propose a novel Pixel-wise Contextual Attention network, which is referred as PiCANet, to learn these informative contextual regions for each image pixel.
It significantly improves the soft attention model \cite{bahdanau2015attention} by generating contextual attention for every pixel
, which is a genuine novel idea for the whole neural network community. 
Specifically, as shown in Figure~\ref{figure1}, the proposed PiCANet learns to generate soft attention over the context regions for each pixel, where the attention weights indicate how relevant each context location is w.r.t. the referred pixel.
The features from the context regions are then weighted and aggregated to obtain an attended contextual feature, which only considers informative context locations while ignores detrimental ones for each pixel.
As a result, the proposed PiCANets can facilitate the saliency detection task significantly.

To incorporate contexts with different scopes, we formulate the PiCANet in two forms: global and local PiCANets, to selectively integrate global context and local context, respectively.
Furthermore, our implementations of the PiCANets are fully differentiable. Thus they can be flexibly embedded into ConvNets and enable joint training.

We hierarchically embed global and local PiCANets into a U-Net architecture \cite{ronneberger2015unet}, which is an encoder-decoder convolutional network with skip connections, to detect salient objects.
In the decoder, we progressively employ several global and local PiCANets on multiscale feature maps.
Thus, we construct the attended contextual features from the global view to local contexts, from coarse scale to fine scales, and use them to enhance the convolutional features to facilitate saliency inference at each pixel.
Figure~\ref{figure1} shows some examples of the learned attention maps.
For each pixel (the red and the blue dots), the learned global attention shown in  Figure~\ref{figure1}(b) can attend to backgrounds for foreground objects and vice verse, which exactly matches the global contrast mechanism.
While the learned local attention shown in  Figure~\ref{figure1}(c) can attend to regions that have the similar appearance with the referred pixel in its local context to make the saliency map more homogeneous.

Our contributions can be summarized as follows:

1. We propose the novel PiCANet to 
generate attention over the context regions for each pixel. Consequently, informative contextual features can be obtained to facilitate the final decision. 
Furthermore,
we formulate PiCANet in both global and local forms to attend to global and local contexts, respectively, and with full differentiability to enable joint training with ConvNets.
%

2.  We propose a novel saliency detection model by embedding PiCANets into a U-Net architecture. PiCANets are used to hierarchically incorporate the attended global context and multiscale local contexts, which can effectively improve saliency detection performance.

3.  Extensive experimental results on six benchmark datasets demonstrate the effectiveness of the proposed PiCANets and the saliency model when compared with other state-of-the-art models.
We also present in-depth analyses and explain why the proposed PiCANets perform well.

\section{Related Work}

\paragraph{Attention networks.}

Recently, attention models are introduced into neural networks to mimic the visual attention mechanism of focusing on informative regions in visual scenes.
Mnih \etal \cite{mnih2014attention} propose a recurrent attention model with hard alignment.
However, it is difficult to train such hard attention models.
Subsequently, Bahdanau \etal \cite{bahdanau2015attention} develop an attention model with differentiable soft alignments for machine translation.
In recent years, attention models have been applied to several vision tasks.
Xu \etal \cite{xu2015show} use an recurrent attention model for image caption to align words with image regions.
In \cite{sermanet2014attention}, Sermanet \etal adopt a recurrent attention model for fine-grained classification via attending to discriminative regions.
In addition, attention models are introduced for visual question answering to attend to question-related image regions \cite{xu2016ask,yang2016stacked}.
Li \etal \cite{li2017attentive} utilize attention to attend to the global context to guide object detection. These works demonstrate that attention models can be significantly helpful for computer vision tasks via attending to informative contexts. However, existing approaches only consider generating one global contextual attention map at one time, which we refer as the \emph{image-wise contextual attention}.
These models limit the application of attention networks in convolutional nets, especially for pixel-wise tasks, since different pixels have different informative context regions.
In \cite{chen2016attention2scale}, Chen \etal generate attention weights for each pixel for semantic segmentation. Nevertheless, this method uses attention to select adaptive scales on multiscale features for each pixel, which we refer as the \emph{pixel-wise scale attention}. In contrast, our proposed PiCANet generates attention for context regions of each pixel.

\vspace{-3mm}
\paragraph{Saliency detection.}

Traditional saliency models mainly rely on various saliency cues to detect salient objects, including local contrast \cite{klein2011center}, global contrast \cite{cheng2015global}, and background prior \cite{yang2013gbmr}. Lately, with the utilization of CNNs, many work have achieved promising results on saliency detection. Next, we briefly review these models.

Liu \etal \cite{liu2015mrcnn} and Li and Yu \cite{li2015mdf} adopt CNNs to extract multiscale contextual features on multiscale image regions to infer saliency for each pixel and each superpixel, respectively. Similarly, Zhao \etal \cite{zhao2015mcdl} use CNNs on both global and local contexts. In \cite{li2016dcl}, an FCN based saliency model and a multiscale image region based saliency model are combined. Wang \etal \cite{wang2016rfcn} recurrently adopt an FCN to refine saliency maps progressively. Liu and Han \cite{liu2016dhsnet} use a U-Net based network to hierarchically predict and refine saliency maps from the global view to finer local views. Similarly, Luo \etal \cite{luo2017nldf} and Zhang \etal \cite{Zhang2017amulet} also utilize U-Net based models to incorporate multi-level contexts to detect salient objects. Wang \etal \cite{Wang2017srm} also use several stages to progressively refine saliency maps by combining local and global context information. In \cite{hou2017dss}, short connections are introduced into the multi-scale side outputs within the HED network \cite{xie2015hed} to improve saliency detection performance. Hu \etal \cite{hu2017dls} propose to adopt a level sets based loss to train their saliency detection network and use guided super-pixel filtering to refine saliency maps.

Although existing DNN based models incorporate various contexts for saliency detection, these methods all use context regions holistically. Typically, the work in \cite{liu2016dhsnet,luo2017nldf,Zhang2017amulet}, which have similar U-Net architectures with the one we use in this paper, incorporate multiscale contexts via diverse network architectures which indiscriminately integrate the information from their receptive field. In contrast, we use the proposed PiCANets to only selectively attend to informative context locations. In \cite{kuen2016recurrent}, authors use a recurrent attention model to select local regions to refine their saliency maps. However, they adopt the spatial transformer attention network \cite{jaderberg2015stn} to select one refining region at each time step, where their model still falls into the \emph{image-wise attention} category. In contrast, our PiCANets can generate soft contextual attention for each pixel.

\section{Pixel-wise Contextual Attention Network}

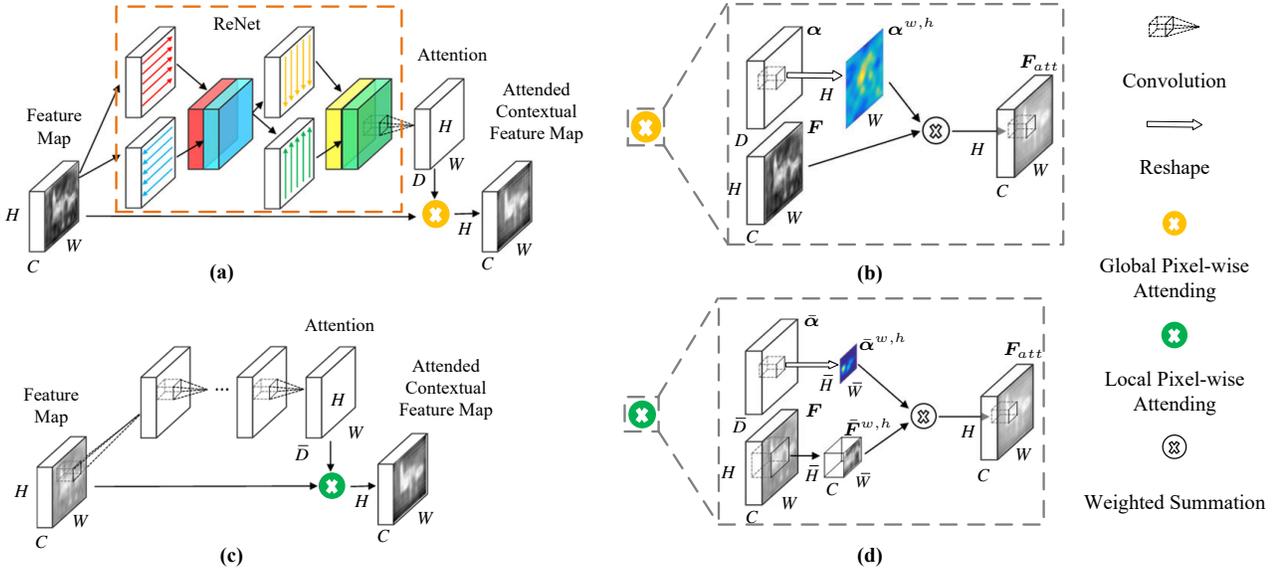
\begin{figure*}[!t]
  \graphicspath{{Figures/PiCANet/}}
  \centering
  \begin{overpic}[width=1\linewidth]{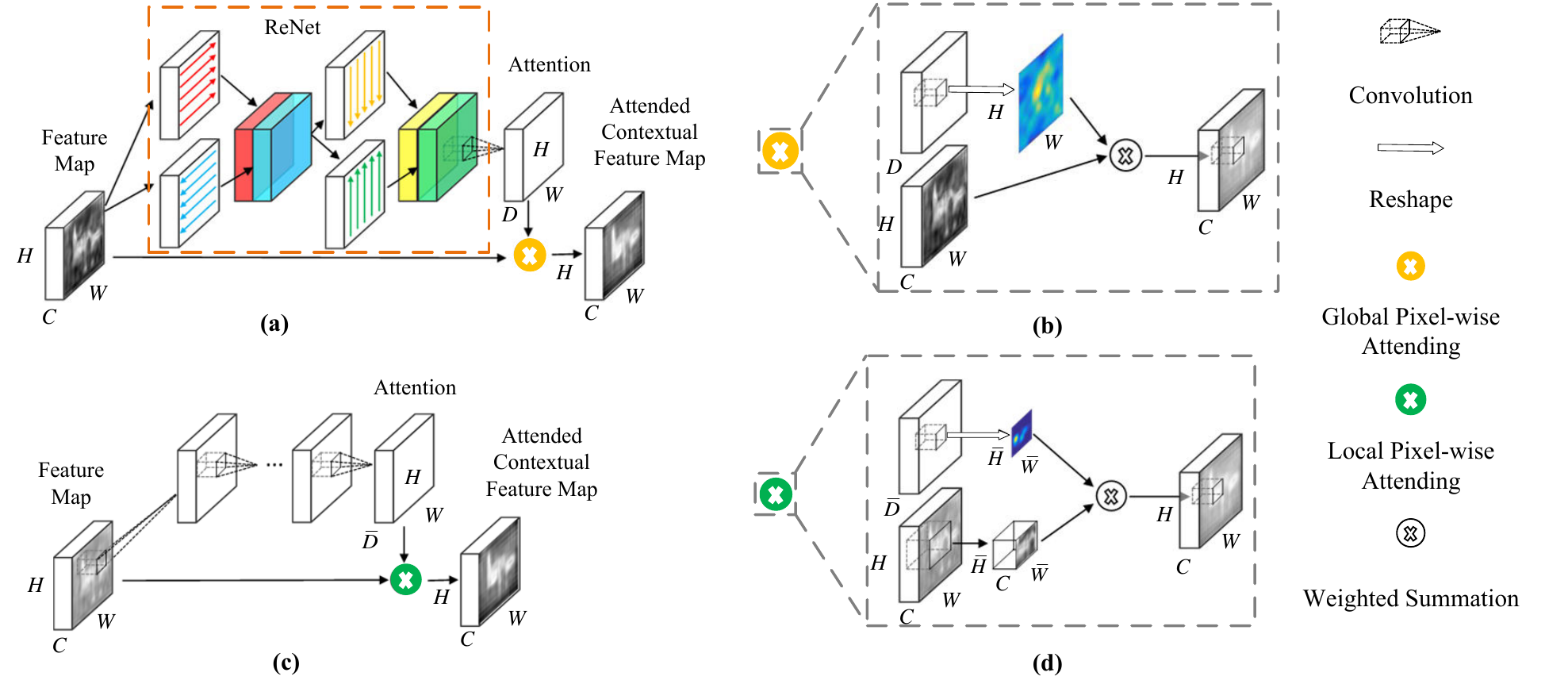}
  \put(62,41){\scriptsize $\bm{\alpha}$}
  \put(68.2,41){\scriptsize $\bm{\alpha}^{w,h}$}
  \put(62,33.5){\scriptsize $\bm{F}$}
  \put(78.2,38.5){\scriptsize $\bm{F}_{att}$}
  \put(61.8,18.8){\scriptsize $\bar{\bm{\alpha}}$}
  \put(66,17){\scriptsize $\bar{\bm{\alpha}}^{w,h}$}
  \put(61.9,11.9){\scriptsize $\bm{F}$}
  \put(64.8,10.5){\scriptsize $\bar{\bm{F}}^{w,h}$}
  \put(77,16.6){\scriptsize $\bm{F}_{att}$}
  \end{overpic}
  \caption{(a) Architecture of the proposed global PiCANet. (b) Illustration of the detailed global attending operation. (c) Architecture of the proposed local PiCANet. (d) Illustration of the detailed local attending operation.}
  \label{PiCANetFig}
  \vspace{-0.3cm}
\end{figure*}

The proposed PiCANet aims at generating an attention map at each pixel over its context region and constructing an attended contextual feature to enhance the feature representability of Convnets. Given a convolutional (Conv) feature map $\bm{F}\in{\mathbb{R}^{W\times{H\times C}}}$, where $W$, $H$, $C$ denote its width, height and number of channels, respectively, we propose two pixel-wise attention modes: global attention and local attention. For each location $(w,h)$ in $\bm{F}$, the former generates attention over the whole feature map $\bm{F}$, while the latter works on a local region centered at $(w,h)$.

\subsection{Global PiCANet}

For the global attention, we show the network architecture in Figure~\ref{PiCANetFig}(a). Since we tend to generate attention over the global context for each pixel, we need to make each pixel be able to ``see'' the overall feature map $\bm{F}$ first. To this end, one can use various network architectures whose receptive field is the whole image, \eg, a fully connected layer. Here we employ a more effective and efficient ReNet model \cite{visin2015renet}, which uses four recurrent neural networks to sweep an image both horizontally and vertically along both directions, to incorporate the global context. Specifically, as shown in the orange dashed box in Figure~\ref{PiCANetFig}(a), a bidirectional LSTM (biLSTM) \cite{graves2013bilstm} is first deployed along each row of $\bm{F}$, then the two hidden states of each pixel are concatenated, making each pixel memorize both its left and right contexts. Next, another biLSTM is deployed along each column of the obtained feature map, so that each pixel can memorize both its top and bottom contexts. By alternately scanning horizontally and vertically, the contexts from four directions can be blended, which propagate the information of each pixel to all other pixels. Thus, global context is efficiently incorporated at each pixel.

Next, we use a vanilla Conv layer to transform the ReNet feature map to $D$ channels, where $D=W\times H$. Then, at each pixel $(w,h)$, the obtained feature vector, which is denoted as $\bm{x}^{w,h}$, is normalized via a softmax function to generate the global attention weights $\bm{\alpha}^{w,h}$:
\begin{equation} \label{attSoftmax}
\alpha_{i}^{w,h}=\frac{\exp{(x_{i}^{w,h})}}{\sum_{j=1}^D\exp{(x_{j}^{w,h})}},
\end{equation}
where $i\in{\{1,\cdots,D\}}$, $\bm{x}^{w,h},\bm{\alpha}^{w,h}\in{\mathbb{R}^D}$, and $\alpha_{i}^{w,h}$ corresponds to the contextual relevance at the $i^{th}$ context location $(W_i, H_i)$ w.r.t. the referred pixel $(w,h)$.

Finally, as shown in Figure~\ref{PiCANetFig}(b), for the pixel $(w,h)$, the features at all locations in $\bm{F}$ are weighted summed by $\bm{\alpha}^{w,h}$ to construct the attended contextual feature $\bm{F}_{att}$:
\begin{equation} \label{globalAtt}
\bm{F}_{att}^{w,h}=\sum_{i=1}^D\alpha_{i}^{w,h}\bm{f}_{i},
\end{equation}
where $\bm{f}_{i}\in{\mathbb{R}^C}$ is the Conv feature at $(W_i, H_i)$ in $\bm{F}$ and $\bm{F}_{att}$ has the same size with $\bm{F}$.

\subsection{Local PiCANet}

As for the local attention, at each pixel $(w,h)$, we only perform the attending operation on a local neighboring context region centered at $(w,h)$, which forms a local feature cube $\bar{\bm{F}}^{w,h}\in{\mathbb{R}^{\bar{W}\times{\bar{H}\times C}}}$, with the width $\bar{W}$ and the height $\bar{H}$. The network architecture is shown in Figure~\ref{PiCANetFig}(c). Again, we first need each pixel to ``see'' the $\bar{W}\times\bar{H}$ context region. We simply use Conv layers to achieve this purpose. Specifically, we deploy several Conv layers on $\bm{F}$ to make their receptive field achieve the size of $\bar{W}\times\bar{H}$. Then, as the same as global PiCANet, a Conv layer is used to transform the resultant feature map to $\bar{D}=\bar{W}\times\bar{H}$ channels. Next, the local attention weights $\bar{\bm{\alpha}}^{w,h}$ are also generated by the softmax normalization (similar to \eqref{attSoftmax}). Finally, as shown in Figure~\ref{PiCANetFig}(d), for pixel $(w,h)$, the features in $\bar{\bm{F}}^{w,h}$ are weighted summed by $\bar{\bm{\alpha}}^{w,h}$ to obtain $\bm{F}_{att}$:
\begin{equation} \label{localAtt}
\bm{F}_{att}^{w,h}=\sum_{i=1}^{\bar{D}}{{\bar{\alpha}}_i^{w,h}\bar{\bm{f}}_{i}^{w,h}}.
\end{equation}

\subsection{Effective and Efficient Implementation}

For computational efficiency, the attending operation for all pixels can be implemented simultaneously by a convolution-like way. We can also adopt the hole algorithm \cite{chen2016deeplab} in the attending operation, which supports sparsely sampling feature maps by using dilated convolution. Thus, we can use a small $D$ or $\bar{D}$ with dilation to attend to large context regions to make PiCANets more efficient. The gradients of the PiCANets can be easily calculated, making end-to-end training feasible via the back-propagation algorithm \cite{rumelhart1988bp}. We can also use a batch normalization (BN) \cite{ioffe2015bn} layer before softmax normalization to make the network training more effective.
\section{Salient Object Detection using PiCANets}

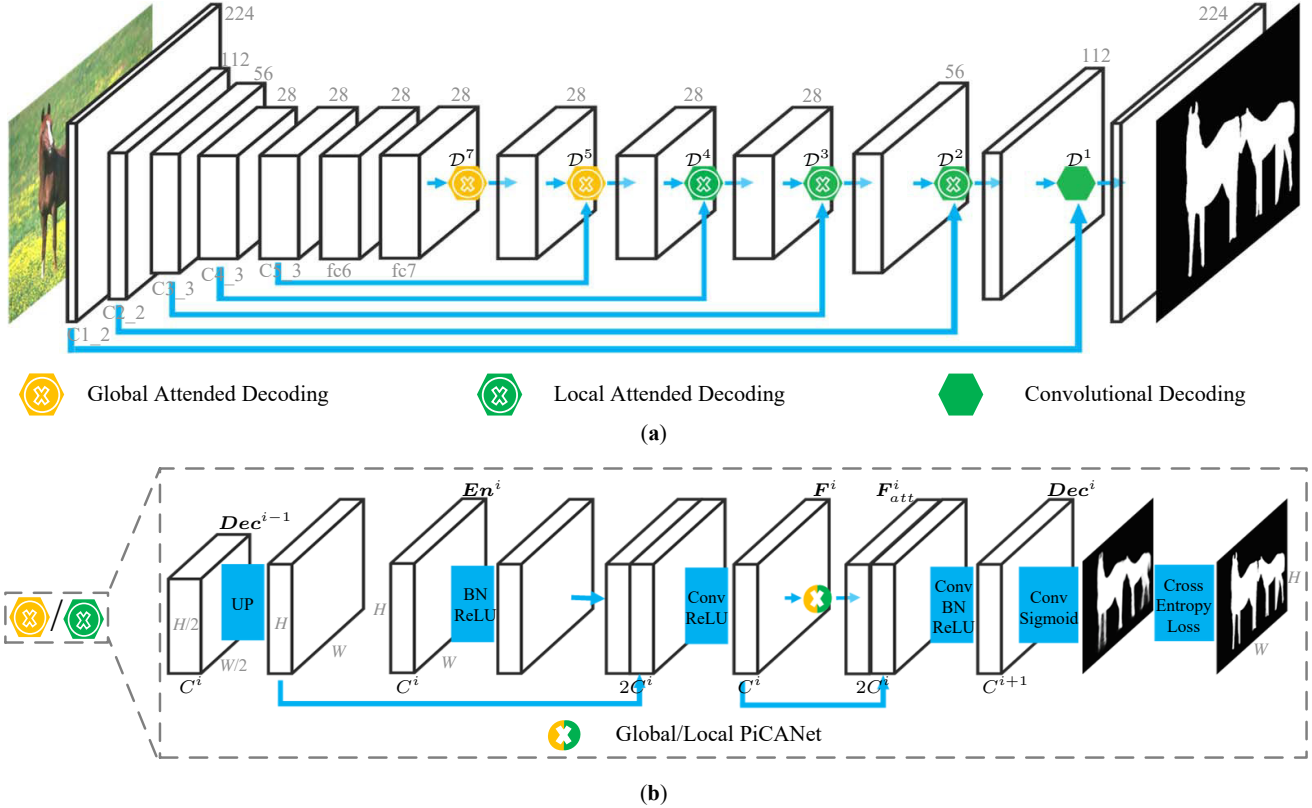
\begin{figure*}[!t]
  \graphicspath{{Figures/PiCANet/}}
  \centering
  \begin{overpic}[width=1\linewidth]{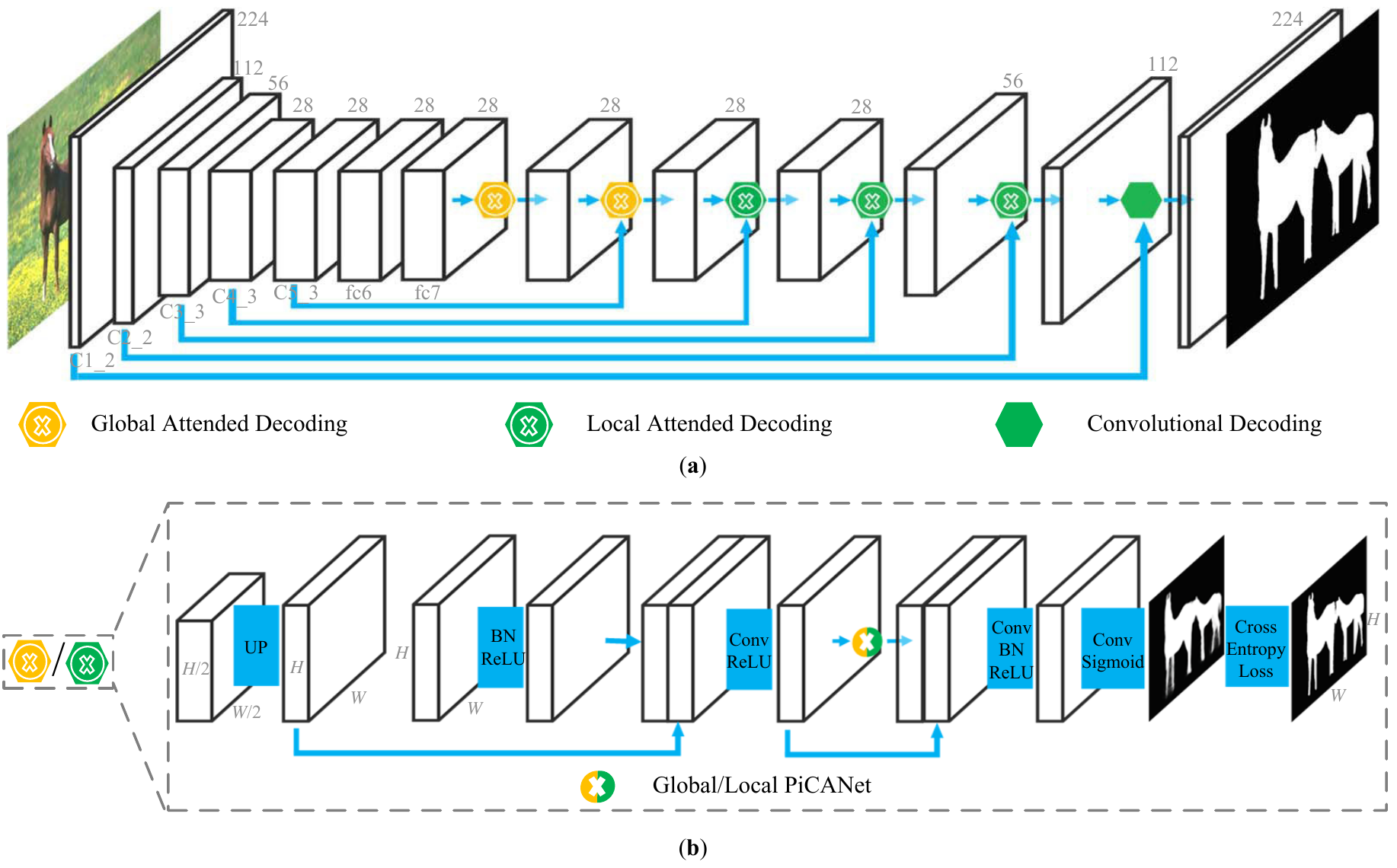}
  \put(34,49.2){\scriptsize $\mathcal D^7$}
  \put(43,49.2){\scriptsize $\mathcal D^5$}
  \put(52,49.2){\scriptsize $\mathcal D^4$}
  \put(61,49.2){\scriptsize $\mathcal D^3$}
  \put(71.2,49.2){\scriptsize $\mathcal D^2$}
  \put(81,49.2){\scriptsize $\mathcal D^1$}
  \put(13.5,9){\scriptsize $C^i$}
  \put(30,9){\scriptsize $C^i$}
  \put(47,9){\scriptsize $2C^i$}
  \put(56,9){\scriptsize $C^i$}
  \put(65,9){\scriptsize $2C^i$}
  \put(74.6,9){\scriptsize $C^{i+1}$}
  \put(16.5,21.5){\scriptsize $\bm{Dec}^{i-1}$}
  \put(35,24){\scriptsize $\bm{En}^i$}
  \put(61.7,24){\scriptsize $\bm{F}^i$}
  \put(66.5,24){\scriptsize $\bm{F}_{att}^i$}
  \put(79.5,24){\scriptsize $\bm{Dec}^i$}
  \end{overpic}
  \caption{(a) Architecture of our saliency network with the VGG 16-layer backbone. We only show the skip-connected encoder layers   of the VGG network. ``C'' means ``convolution'' while $\mathcal D^*$ indicates a decoding module. The spatial sizes are marked over the cuboids which represent the feature maps. (b) Illustration of an attended decoding module. $\bm{En}^i$ denotes a convolutional feature map from the encoder network. $\bm{Dec}^*$ denotes a decoding feature map. $\bm{F}^i$ denotes a fusion feature map and $\bm{F}_{att}^i$ denotes its attended contextual feature map. ``UP'' denotes upsampling. Some important spatial sizes and channel numbers are also marked.}
  \label{SONetFig}
  \vspace{-0.3cm}
\end{figure*}

In this section, we elaborate our network architecture which adopts PiCANets hierarchically for salient object detection. The whole network is based on a U-Net \cite{ronneberger2015unet} architecture as shown in Figure~\ref{SONetFig}(a). However, different from \cite{ronneberger2015unet}, the encoder of our U-Net is an FCN with the hole algorithm \cite{chen2016deeplab} to keep the resolutions of feature maps. The decoder follows the idea of U-Net to use skip connections and with our proposed global and local PiCANets embedded.

Considering the global PiCANet requires the input feature map to have a fixed size, we set input images to have a fixed size of $224\times 224$. The encoder part is an FCN with a pretrained backbone network, \eg, the VGG \cite{simonyan2014vgg} network or a ResNet \cite{he2016resnet}. We take the VGG 16-layer network as an example, which contains 13 Conv layers, 5 max-pooling layers, and 2 fully connected layers. As shown in Figure~\ref{SONetFig}(a), in order to preserve relative large spatial sizes in higher layers for accurate saliency detection, we modify the pooling strides of the pool4 and pool5 layers to be 1 and adopt the hole algorithm \cite{chen2016deeplab} to introduce dilation of 2 for the conv5 layers. We also follow \cite{chen2016deeplab} to transform the last 2 fully connected layers to Conv layers. Specifically, we use 1024 $3\times 3$ kernels with dilation of 12 for the fc6 layer and 1024 $1\times 1$ kernels for the fc7 layer. Thus, the stride of the whole encoder network is reduced to 8, and the spatial size of the final feature map is $28\times 28$.

Next, we elaborate our decoder part. As shown in Figure~\ref{SONetFig}(a), the decoder network has 6 decoding modules, named $\mathcal D^7,\mathcal D^5,\mathcal D^4,\cdots,\mathcal D^1$. As shown in Figure~\ref{SONetFig}(b), in $\mathcal D^i$, where $i\in{\{7,5,4,\cdots,1\}}$, we usually generate a decoding feature map $\bm{Dec}^i$ by fusing an intermediate encoder feature map $\bm{En}^i$ with the size of $W\times{H\times C^i}$ and the preceding decoding feature map $\bm{Dec}^{i-1}$ with the size of $W/2\times{H/2\times C^i}$. $\bm{En}^i$ is the Conv feature map before the ReLU activation of the $i^{th}$ Conv block in the VGG encoder part, and they are marked in Figure~\ref{SONetFig}(a). We first use a BN layer and the ReLU activation on $\bm{En}^i$. At the same time, we upsample $\bm{Dec}^{i-1}$ to have the spatial size of $W\times H$ by using a deconvolutional layer with bilinear interpolation. Next, we concatenate these two feature maps and fuse them into a feature map $\bm{F}^i$ with $C^i$ channels by using a Conv and a ReLU layer. Then we utilize a global or a local PiCANet on $\bm{F}^i$ to obtain its attended contextual feature map $\bm{F}_{att}^i$. Finally we fuse $\bm{F}^i$ and $\bm{F}_{att}^i$ into $\bm{Dec}^i$ with the size $W\times{H\times C^{i+1}}$, via a Conv layer, a BN layer, and a ReLU layer. We also adopt deep supervision to facilitate the network training. Specifically, in each $\mathcal D^i$, we use a Conv layer with sigmoid activation on $\bm{Dec}^i$ to generate a saliency map with size $W\times H$, then the resized ground truth saliency map is used to supervise the network training based on the average cross-entropy loss.

In each $\mathcal D^i$, we set $C^i$ to be the same as the channel number of the $i^{th}$ Conv block in the encoder network. We adopt global PiCANets in $\mathcal D^7$ and $\mathcal D^5$ and local PiCANets in the next three decoding modules. For $\mathcal D^1$, we simply fuse $\bm{En}^1$ and $\bm{Dec}^2$ into $\bm{Dec}^1$ with simple Conv layers for computational efficiency. The influence of different embedding choices of global and local PiCANets is shown in Section~\ref{sec:ablation}.

\section{Experiments}
\subsection{Datasets}

We use six widely used saliency benchmark datasets to evaluate our method. \textbf{SOD} \cite{movahedi2010sod} contains 300 images with complex backgrounds and multiple foreground objects. \textbf{ECSSD} \cite{yan2013hs} has 1,000 semantically meaningful and complex images. The \textbf{PASCAL-S} \cite{li2014secrets} dataset consists of 850 images selected from the PASCAL VOC 2010 segmentation dataset. \textbf{DUT-O} \cite{yang2013gbmr} includes 5,168 challenging images, each of which usually has complicated background and one or two foreground objects. \textbf{HKU-IS} \cite{li2015mdf} contains 4,447 images with low color contrast and multiple foreground objects in each image. The last one is the \textbf{DUTS} \cite{wang2017duts} dataset, which is currently the largest salient object detection benchmark dataset. It contains 10,553 images in the training set, \ie DUTS-TR, and 5,019 images in the test set, \ie DUTS-TE. Most of the images have challenging scenarios for saliency detection.
\subsection{Evaluation Metrics}

We adopt four evaluation metrics to evaluate our model. The first one is the precision-recall (PR) curve. Specifically, saliency maps are first binarized and then compared with the ground truth under varying thresholds, thus obtaining a series of precision-recall value pairs to draw the PR curve.

The second metric is the F-measure score which comprehensively considers both precision and recall:
\begin{equation} \label{fmeasure}
F_{\beta}=\frac{(1+\beta^2)Precision\times Recall}{\beta^2 Precision+Recall},
\end{equation}
where we set $\beta^2$ to 0.3 as suggested in previous work.

However, as demonstrated in \cite{margolin2014evaluate}, traditional evaluation metrics easily suffer from the interpolation flaw, dependency flaw, and equal-importance flaw, leading to unfair comparison. Thus the authors propose the \emph{weighted} F-measure score $F_{\beta}^{\omega}$ to address these drawbacks. We also follow \cite{gong2015saliency,tu2016real,hu2017dls} to adopt it as one of our metrics with the default settings in \cite{margolin2014evaluate}.

The fourth metric we use is the Mean Absolute Error (MAE). It computes the average absolute per-pixel difference between predicted saliency maps and corresponding ground truth saliency maps.
\subsection{Implementation Details}

\paragraph{Network structure.}
In the decoding modules, all of the convolutional kernels in Figure~\ref{SONetFig}(b) are set to $1\times 1$. In each global PiCANet, we use 256 hidden neurons for the ReNet, then we use a $1\times 1$ Conv layer to generate $D=100$ dimensional attention weights, which can be reshaped to $10\times 10$ attention maps. In its attending operation, we use $dilation=3$ to attend to the $28\times 28$ global context. In each local PiCANet, we first use a $7\times 7$ Conv layer with $dilation=2$, zero padding, and ReLU activation to generate an intermediate feature map with 128 channels. Then we adopt a $1\times 1$ Conv layer to generate $\bar{D}=49$ dimensional attention weights, from which $7\times 7$ attention maps can be obtained. Then we utilize these local attention maps to attend to $13\times 13$ local context regions with $dilation=2$ and zero padding.

\paragraph{Training and testing.}
We follow \cite{Wang2017srm} and the suggestion in \cite{wang2017duts} to use the DUTS-TR set as our training set. For data augmentation, we simply resize each image to $256\times 256$ with random mirror-flipping and randomly crop $224\times 224$ image regions for training. The whole network is trained end-to-end using stochastic gradient descent (SGD) with momentum. Since deep supervision is adopted in each decoding module, we empirically weight the losses in $\mathcal D^7,\mathcal D^5,\mathcal D^4,\cdots,\mathcal D^1$ by 0.5, 0.5, 0.5, 0.8, 0.8, and 1, respectively, without further tuning. We train the decoder part from scratch with a learning rate of 0.01 and finetune the encoder with a 0.1 times smaller learning rate. We set the batchsize to 10, the maximum iteration step to 20,000, and decay the learning rates by a factor of 0.1 every 7,000 steps. The momentum and the weight decay are set to 0.9 and 0.0005, respectively.

We implement our model based on the Caffe \cite{jia2014caffe} library. A GTX Titan X GPU is used for acceleration. When testing, each image is simply resized to $224\times 224$ and then fed into the network to obtain its saliency map. The testing process only costs 0.178s for each image when using the VGG 16-layer backbone. Our code will be released.
\subsection{Ablation Study}\label{sec:ablation}

\paragraph{Effectiveness of the proposed PiCANets.}

\begin{table} [!t]
\begin{center}
\caption{Quantitative results of different settings of our model and baseline models. ``MP'' and ``AP'' mean max-pooling and average pooling, respectively. ``+75G432LP'' means using \textbf{G}lobal PiCANets in $\mathcal D^7$ and $\mathcal D^5$, and \textbf{L}ocal \textbf{P}iCANets in $\mathcal D^4$, $\mathcal D^3$, $\mathcal D^2$. Other settings can be inferred similarly. \blu{Blue} indicates the best performance.}
\vspace{1mm}
\label{ablationTab}
\footnotesize
\begin{tabular}{@{}lccccccc@{}}
\toprule
\multirow{2}{*}{Settings} & \multicolumn{3}{c}{DUT-O \cite{yang2013gbmr}}  && \multicolumn{3}{c}{DUTS-TE \cite{wang2017duts}} \\
\cmidrule{2-4} \cmidrule{6-8}
                          &  $F_{\beta}$   &  $F_{\beta}^{\omega}$  &   MAE    && $F_{\beta}$   &  $F_{\beta}^{\omega}$  &   MAE  \\ \midrule
U-Net \cite{ronneberger2015unet} & 0.761   & 0.651                  & 0.073    && 0.819         & 0.715                  & 0.060  \\ \midrule
+75GP                     & 0.778          & 0.662                  & 0.071    && 0.834         & 0.724                  & 0.057  \\
+75G432LP                 &\blu{0.794}     &\blu{0.691}             & 0.068    &&\blu{0.851}    &\blu{0.748}             & 0.054  \\ \midrule
+MP                       & 0.780          & 0.671                  & 0.070    && 0.833         & 0.727                  & 0.057  \\
+AP                       & 0.778          & 0.670                  & 0.069    && 0.831         & 0.724                  & 0.056  \\ \midrule
+75432LP                  & 0.787          & 0.680                  & 0.069    && 0.842         & 0.738                  & 0.055  \\
+7G5432LP                 & 0.792          & 0.690                  & 0.069    && 0.849         & 0.744                  & 0.054  \\
+754G32LP                 & 0.794          & 0.688                  &\blu{0.065} && 0.850         & 0.747                &\blu{0.053}  \\
\bottomrule
\end{tabular}
\vspace{-0.7cm}
\end{center}{}
\end{table}

To demonstrate the effectiveness of the proposed PiCANets, we show quantitative comparison results of our model against baseline models on two challenging datasets in Table~\ref{ablationTab}. ``U-Net'' is the baseline network without PiCANets. ``+75GP'' means we only embed two global PiCANets into $\mathcal D^7$ and $\mathcal D^5$, while ``+75G432LP'' means we embed global PiCANets into $\mathcal D^7$ and $\mathcal D^5$, and local PiCANets into $\mathcal D^4$, $\mathcal D^3$, $\mathcal D^2$. The comparison results show that when we gradually use PiCANets to incorporate global and multiscale local contexts selectively, the model performance can be progressively boosted. A more detailed ablation study of progressively embedding PiCANets in each decoding module is given in the supplementary material.

\begin{figure*}[!t]
  \graphicspath{{Figures/PR_curves/}}
  \centering
  \includegraphics[width=1\linewidth]{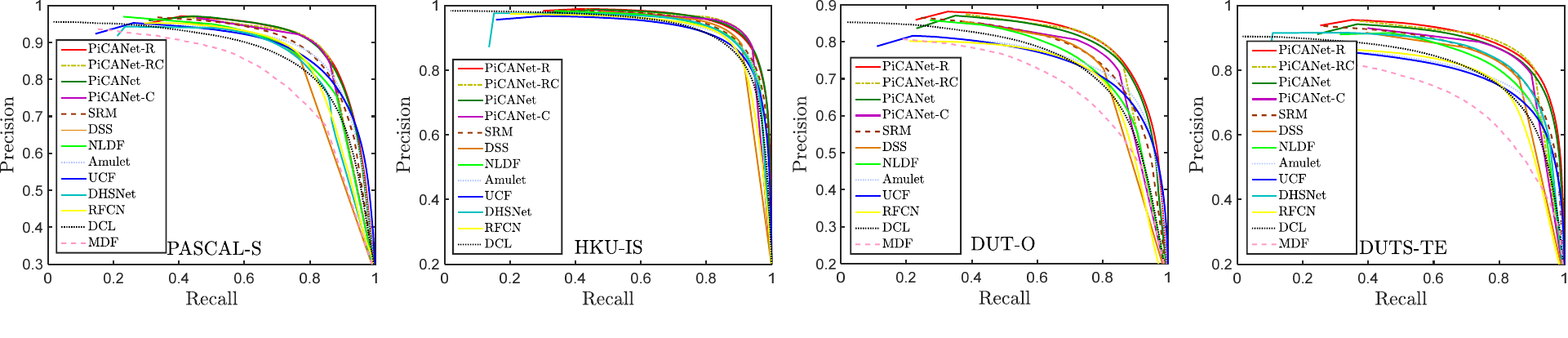}
  \caption{Comparison on four large datasets in terms of the PR curve.}
  \label{prcurve}
  \vspace{-0.1cm}
\end{figure*}

For a fair comparison, we also adopt max-pooling (MP) and average-pooling (AP) to incorporate these contexts. Table~\ref{ablationTab} shows that although using these non-parametric pooling schemes to incorporate global and local contexts can bring performance gains, using our proposed PiCANets to select informative contexts is a much better way.

\begin{figure}[!ht]
  \graphicspath{{Figures/baselinCmp/}}
  \centering
  \includegraphics[width=1\linewidth]{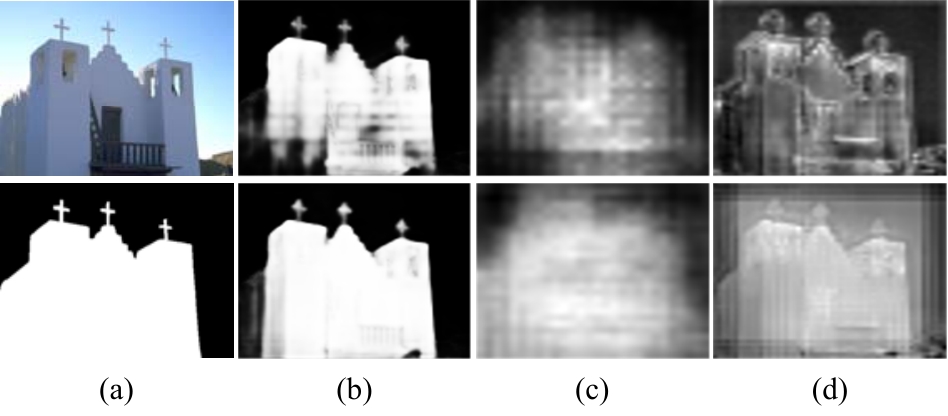}
  \caption{Visual comparison of our model against the baseline U-Net. (a) An image and its ground truth. (b) Saliency maps of the baseline U-Net (top) and our model (bottom). (c) $\bm{F}^5$ (top) and $\bm{F}_{att}^5$ (bottom). (d) $\bm{F}^2$ (top) and $\bm{F}_{att}^2$ (bottom).}
  \label{baselinCmp}
  \vspace{-0.3cm}
\end{figure}

We also show visual comparison results to demonstrate the effectiveness of the proposed PiCANets. In Figure \ref{baselinCmp}(a) we show an image and its ground truth saliency map while (b) shows the predicted saliency maps of the baseline U-Net (top) and our model (bottom). We can see that our saliency model can obtain more uniformly highlighted saliency map with the help of PiCANets. In Figure \ref{baselinCmp}(c), we show a comparison of the Conv feature map $\bm{F}^5$ (top) against the attended contextual feature map $\bm{F}_{att}^5$ (bottom) with the global PiCANet. While (d) shows $\bm{F}^2$ (top) and $\bm{F}_{att}^2$ (bottom) with the local PiCANet. We can see that the global PiCANet in $\mathcal D^5$ helps to better discriminate the foreground object from backgrounds, while the local PiCANet in $\mathcal D^2$ enhances the feature map to be more homogenous, which makes the whole foreground object highlighted more uniformly.

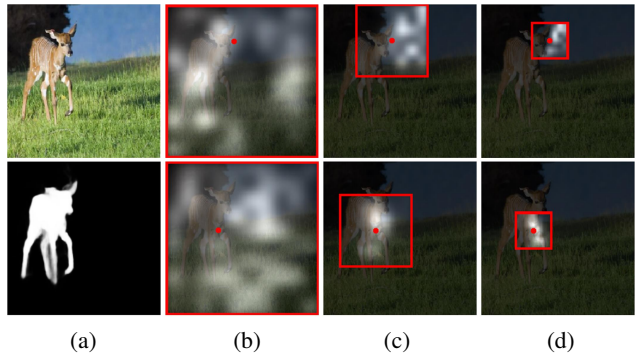
\begin{figure}[!ht]
  \graphicspath{{Figures/attention/}}
  \centering
  \begin{overpic}[width=1\linewidth]{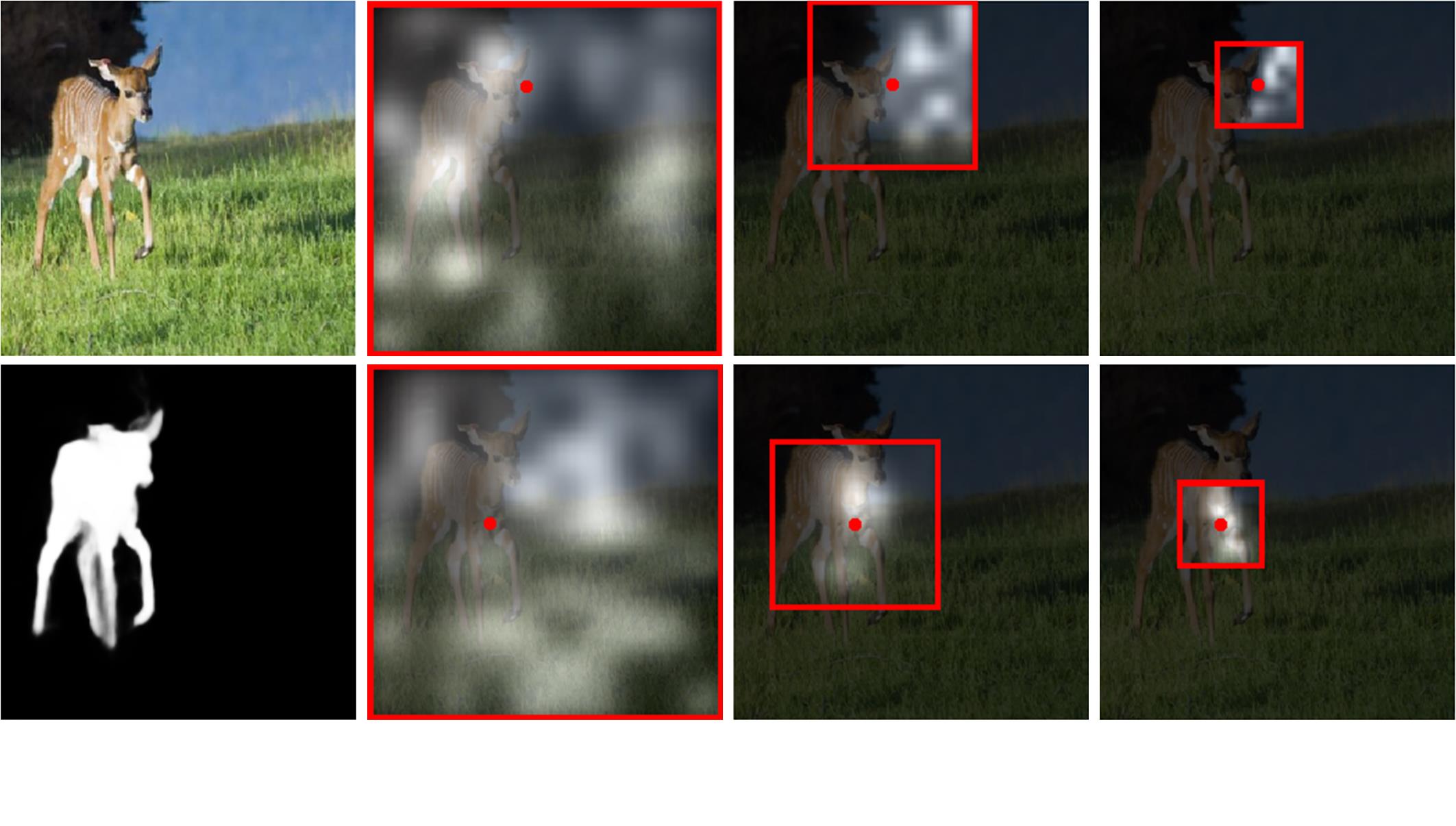}
  \put(10,2){\small (a)}
  \put(36,2){\small (b)}
  \put(60,2){\small (c)}
  \put(86,2){\small (d)}
  \end{overpic}
  \caption{Illustration of the learned attention maps of the proposed PiCANet. (a) shows an image and its predicted saliency map of our model. We show the attention maps of two pixels (denoted as red dots. The top row shows a background pixel and the bottom row shows a foreground pixel.) in $\mathcal D^5$ (b), $\mathcal D^4$ (c), and $\mathcal D^3$ (d), respectively. The attended context regions are marked by red rectangles.}
  \label{attention}
  \vspace{-0.4cm}
\end{figure}

To further understand why PiCANets can achieve such improvements, we visualize the learned attention maps of two pixels in one image in Figure~\ref{attention}. In column (b), the top image shows that the global attention of the background pixel mainly attends to the foreground object while the bottom image shows that for the foreground pixel, it mainly attends to the background regions. This observation greatly matches the global contrast mechanism. Thus our global PiCANet can help the network to effectively tell the salient objects from the backgrounds. As for the local attention, since we used fixed attention size ($13\times 13$) for different decoding modules, we can incorporate multiscale attention from coarse to fine, with large contexts to small ones, as shown by red rectangles in Figure~\ref{attention}. The (c) and (d) columns show that local attention mainly attends to homogeneous regions with the referred pixel, thus enhancing the saliency map to be uniform, just as shown in the bottom image in column (a). More visualization can be found in the supplementary material.

\paragraph{Influence of the embedding choice.} We also show comparison results of different embedding choices of our global and local PiCANets in Table~\ref{ablationTab}. It shows that only embedding local PiCANets (``+75432LP'') is inferior. While the results of ``+7G5432LP'' and ``+754G32LP'' are slightly worse than our final choice, \ie ``+75G432LP''. We do not consider to use global PiCANets in other decoding modules since the ReNet is time-consuming for large feature maps.
\subsection{Comparison with State-of-the-arts}

We compare our saliency model against other 9 state-of-the-art models, namely, SRM \cite{Wang2017srm}, DSS \cite{hou2017dss}, NLDF \cite{luo2017nldf}, Amulet \cite{Zhang2017amulet}, UCF \cite{Zhang2017ucf}, DHS \cite{liu2016dhsnet}, RFCN \cite{wang2016rfcn}, DCL \cite{li2016dcl}, and MDF \cite{li2015mdf}.

\begin{table*} [!t]
\begin{center}
\caption{Comparison of different methods on 6 datasets under different settings. \blu{Blue} indicates the best performance under each setting while \red{red} indicates the best performance under all settings. ``-C'', ``-R'', and ``-RC'' means using the CRF postprocessing,  the ResNet50 backbone, and both of them, respectively.}
\vspace{-1mm}
\label{sotaTab}
\footnotesize
\begin{tabular}{@{}L{1.5cm}|C{0.45cm}C{0.45cm}C{0.45cm}|C{0.45cm}C{0.45cm}C{0.45cm}|C{0.45cm}C{0.45cm}C{0.45cm}|C{0.45cm}C{0.45cm}C{0.45cm}|C{0.45cm}C{0.45cm}C{0.45cm}|C{0.45cm}C{0.45cm}C{0.45cm}}
\toprule
Dataset & \multicolumn{3}{c|}{SOD \cite{yang2013gbmr}} & \multicolumn{3}{c|}{ECSSD \cite{yang2013gbmr}} & \multicolumn{3}{c|}{PASCAL-S \cite{li2014secrets}} & \multicolumn{3}{c|}{HKU-IS \cite{li2015mdf}} & \multicolumn{3}{c|}{DUT-O \cite{yang2013gbmr}} & \multicolumn{3}{c}{DUTS-TE \cite{wang2017duts}} \\ \midrule
          Metric  &  $F_{\beta}$   &  $F_{\beta}^{\omega}$  &   MAE   & $F_{\beta}$   &  $F_{\beta}^{\omega}$  &   MAE & $F_{\beta}$   &  $F_{\beta}^{\omega}$  &   MAE & $F_{\beta}$   &  $F_{\beta}^{\omega}$  &   MAE & $F_{\beta}$   &  $F_{\beta}^{\omega}$  &   MAE & $F_{\beta}$   &  $F_{\beta}^{\omega}$  &   MAE  \\ \midrule
\multicolumn{19}{c}{VGG-16 \cite{simonyan2014vgg} backbone}\\ \cmidrule{1-19}
MDF \cite{li2015mdf}           &0.760      &0.501      &0.192      &0.832      &0.705      &0.105      &0.782      &0.579      &0.165                                        &-          &-          &-          &0.694      &0.565      &0.092      &0.711      &0.509      &0.114\\
RFCN \cite{wang2016rfcn}       &0.807      &0.592      &0.166      &0.898      &0.727      &0.095      &0.850      &0.671      &0.132                                        &0.898      &0.718      &0.080      &0.738      &0.562      &0.095      &0.783      &0.587      &0.090\\
DHS \cite{liu2016dhsnet}       &0.827      &0.686      &0.133      &0.907      &0.841      &0.060      &0.841      &0.732      &0.111                                        &0.902      &0.806      &0.054      &-          &-          &-          &0.829      &0.698      &0.065\\
UCF \cite{Zhang2017ucf}        &0.803      &0.644      &0.169      &0.911      &0.789      &0.078      &0.846      &0.709      &0.128                                        &0.886      &0.751      &0.074      &0.735      &0.565      &0.132      &0.771      &0.588      &0.117\\
Amulet \cite{Zhang2017amulet}  &0.808      &0.686      &0.145      &0.915      &0.841      &0.059      &0.858      &0.762      &0.103                                        &0.896      &0.813      &0.052      &0.743      &0.626      &0.098      &0.778      &0.657      &0.085\\
NLDF \cite{luo2017nldf}        &0.842      &0.708      &0.130      &0.905      &0.839      &0.063      &0.845      &0.743      &0.112                                        &0.902      &0.838      &0.048      &0.753      &0.634      &0.080      &0.812      &0.710      &0.066\\
PiCANet                        &\blu{0.855}&\blu{0.721}&\blu{0.108}&\blu{0.931}&\blu{0.865}&\blu{0.047}&\blu{0.880}&\blu{0.781}&\blu{0.088}                                &\blu{0.921}&\blu{0.847}&\blu{0.042}&\blu{0.794}&\blu{0.691}&\blu{0.068}&\blu{0.851}&\blu{0.748}&\blu{0.054}\\
\midrule
\multicolumn{19}{c}{VGG-16 \cite{simonyan2014vgg} backbone + CRF \cite{krahenbuhl2011crf}}\\ \cmidrule{1-19}
DCL \cite{li2016dcl}           &0.825      &0.641      &0.198      &0.901      &0.820      &0.075      &0.823      &0.678      &0.189                                        &0.885      &0.736      &0.137      &0.739      &0.575      &0.157      &0.782      &0.606      &0.150\\
DSS \cite{hou2017dss}          &\blu{0.846}&0.718      &0.126      &0.916      &0.871      &0.053      &0.846      &0.751      &0.112                                        &0.911      &0.866      &0.040      &0.771      &0.691      &0.066      &0.825      &0.754      &0.057\\
PiCANet-C                      &0.836      &\blu{0.727}&\blu{0.102}&\blu{0.933}&\blu{0.898}&\blu{0.036}&\blu{0.881}&\blu{0.809}&\blu{0.079}                                   &\blu{0.925}&\blu{0.889}&\blu{0.031}&\blu{0.784}&\blu{0.722}&\blu{0.059}&\blu{0.850}&\blu{0.791}&\blu{0.046}\\
\midrule
\multicolumn{19}{c}{ResNet50 \cite{he2016resnet} backbone}\\ \cmidrule{1-19}
SRM \cite{Wang2017srm}         &0.845      &0.671      &0.132      &0.917      &0.853      &0.054      &0.862      &0.760      &0.098                                        &0.906      &0.836      &0.046      &0.769      &0.658      &0.069      &0.827      &0.722      &0.059\\
PiCANet-R                      &\red{0.858}&\blu{0.723}&\blu{0.109}&\blu{0.935}&\blu{0.867}&\blu{0.047}&\blu{0.881}&\blu{0.780}&\blu{0.087}                                &\blu{0.919}&\blu{0.840}&\blu{0.043}&\blu{0.803}&\blu{0.695}&\blu{0.065}&\blu{0.860}&\blu{0.756}&\blu{0.051}\\
\midrule
\multicolumn{19}{c}{ResNet50 \cite{he2016resnet} backbone + CRF \cite{krahenbuhl2011crf}}\\ \cmidrule{1-19}
PiCANet-RC                     &0.856      &\red{0.742}&\red{0.100}&\red{0.940}&\red{0.908}&\red{0.035}&\red{0.883}&\red{0.812}&\red{0.077}                                   &\red{0.927}&\red{0.890}&\red{0.031}&\red{0.804}&\red{0.743}&\red{0.054}&\red{0.866}&\red{0.811}&\red{0.041}\\
\bottomrule
\end{tabular}
\vspace{-0.4cm}
\end{center}{}
\end{table*}

\begin{figure*}[!ht]
  \graphicspath{{Figures/qualitative/}}
  \centering
  \begin{overpic}[width=1\linewidth]{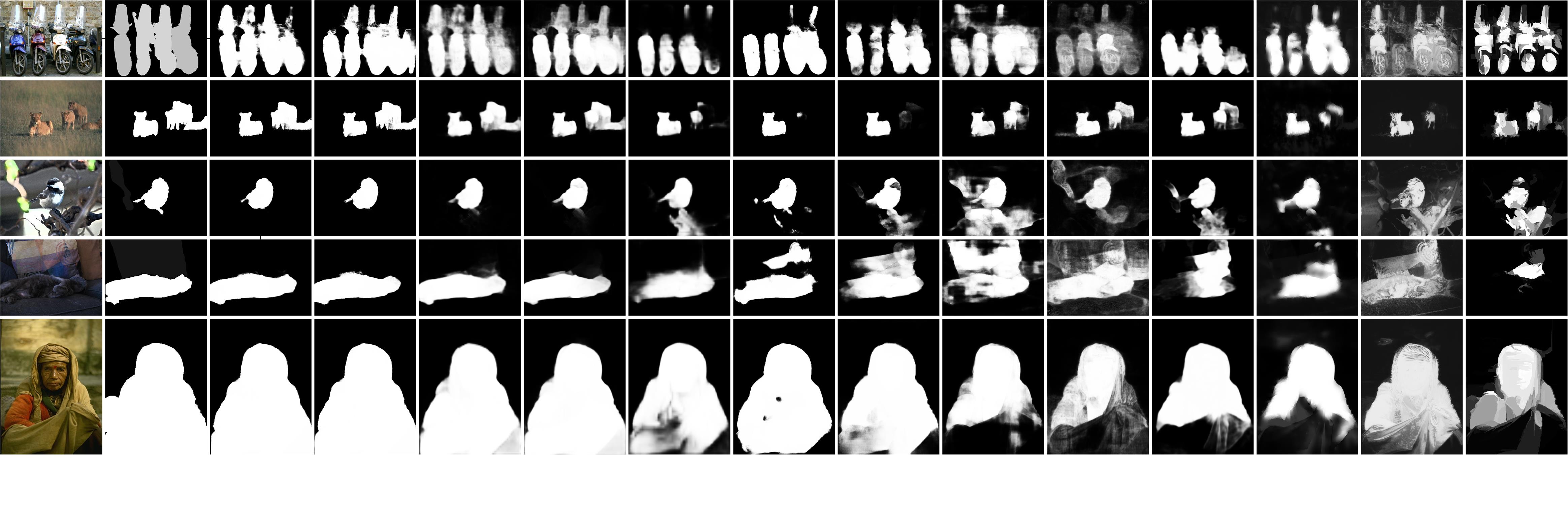}
  \put(1,1.5){\scriptsize Image}
  \put(9,1.5){\scriptsize GT}
  \put(13.8,0.5){\scriptsize \shortstack[c] {PiCANet\\ -RC}}
  \put(20.8,0.5){\scriptsize \shortstack[c] {PiCANet\\ -C}}
  \put(27.5,0.5){\scriptsize \shortstack[c] {PiCANet\\ -R}}
  \put(34.5,1.5){\scriptsize PiCANet}
  \put(40.5,1.5){\scriptsize SRM \cite{Wang2017srm}}
  \put(48,1.5){\scriptsize DSS \cite{hou2017dss}}
  \put(53.5,1.5){\scriptsize NLDF \cite{luo2017nldf}}
  \put(60,1.5){\scriptsize Amulet \cite{Zhang2017amulet}}
  \put(67.3,1.5){\scriptsize UCF \cite{Zhang2017ucf}}
  \put(74,1.5){\scriptsize DHS \cite{liu2016dhsnet}}
  \put(80.2,1.5){\scriptsize RFCN \cite{wang2016rfcn}}
  \put(87.6,1.5){\scriptsize DCL \cite{li2016dcl}}
  \put(94,1.5){\scriptsize MDF \cite{li2015mdf}}
  \end{overpic}
  \caption{Qualitative comparison. (GT: ground truth)}
  \label{visualcmp}
  \vspace{-0.3cm}
\end{figure*}
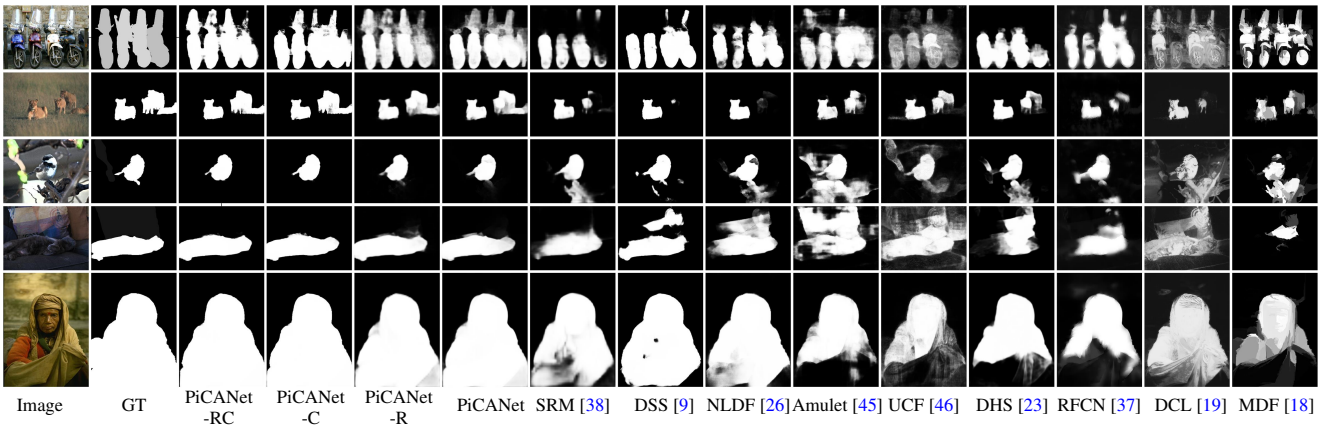

In Table~\ref{sotaTab}, we show the quantitative comparison results. Since \cite{li2016dcl} and \cite{hou2017dss} adopt the fully connected conditional random field (CRF) \cite{krahenbuhl2011crf} as a post-processing technique, while \cite{Wang2017srm} use the ResNet50 \cite{he2016resnet} network as their backbone, for a fair comparison we also adopt them in our model and compare it with other models under different settings. The PR curves on four large datasets are also given in Figure~\ref{prcurve}. We observe that our model consistently performs better than all other models under all settings, especially in terms of the \emph{weighted} F-measure. It is also worth noting that even only use the VGG 16-layer backbone and without any post-processing method, our vanilla PiCANet still performs favorably against all other models. When using both of the CRF post-processing and the ResNet50 backbone, our PiCANet-RC model achieves the best performance and shows significant performance gains over existing methods.

In Figure~\ref{visualcmp}, we show qualitative comparison. We observe that our model can handle various challenging scenarios, including images with complex backgrounds and foregrounds (rows 1, 2, and 3), varying object scales, object touching image boundaries (row 5), object having the similar appearance with the background (row 4). Most importantly, even for the vanilla PiCANet and PiCANet-R which do not use any post-processing methods, they can highlight salient objects more uniformly than other models with the help of PiCANets. More visual comparison results can be found in the supplementary material.

\section{Conclusion}

In this paper, we propose novel PiCANets to selectively attend to global or local contexts and construct informative contextual features for each pixel. We apply PiCANets to detect salient objects in a hierarchical fashion. With the help of attended contexts, our model achieves the best performance on six benchmark datasets. We also provide in-depth analyses of the effectiveness of the PiCANets.

\section*{Acknowledgments}
%
This work is supported in part by the National Science Foundation of China (No. 61473231 and 61522207) and NSF CAREER (No. 1149783).

{\small
\bibliographystyle{ieee}
\bibliography{cvpr18_PiCANet_saliency}
}

\end{document}


\title{PiCANet: Learning Pixel-wise Contextual Attention for Saliency Detection\\
Supplemental Material}

\author{
	Nian Liu$^{1}$
	\hspace{25pt}
	Junwei Han$^{1}$
	\hspace{25pt}
	Ming-Hsuan Yang$^{2,3}$
	\\
	$^1$Northwestern Polytechincal University
	\hspace{8pt}
	$^2$University of California, Merced
	\hspace{8pt}
	$^3$Google Cloud
	\\
	{\tt\small
    \{liunian228, junweihan2010\}@gmail.com
    \hspace{50pt}
    mhyang@ucmerced.edu
    }
}

\maketitle
\thispagestyle{empty}

In this supplementary material, we include more implementation details, more ablation analyses, and more experimental results.

\section{Additional Implementation Details}

\subsection{Using the ResNet50 Backbone}

When using the ResNet50 network \cite{he2016resnet} as the backbone, we modify the $4^{th}$ and the $5^{th}$ residual blocks to have strides of 1 and dilations of 2 and 4, respectively, thus making the encoder have a stride of 8. Then we progressively fuse the feature maps from the $5^{th}$ to the $1^{st}$ Conv blocks in the decoding modules $\mathcal D^5$ to $\mathcal D^1$. We adopt global PiCANets in $\mathcal D^5$ and $\mathcal D^4$, and local PiCANets in the last three modules, respectively. In each decoding module $\mathcal D^i$, we use the final Conv feature map of the $i^{th}$ Conv block in the ResNet50 encoder (\eg res4f and res3d) as the incorporated encoder feature map $\bm{En}^i$ and do not adopt the BN and the ReLU layers on it as shown in Figure 3(b) since the ResNet50 network has already used BN layers after each Conv layer. The final generated saliency map is of size $112\times 112$ since the conv1 layer has a stride of 2.

As the same as when using the VGG-16 backbone, we empirically set the loss weights in $\mathcal D^5,\mathcal D^4,\cdots,\mathcal D^1$ as 0.5, 0.5, 0.8, 0.8, and 1, respectively. The minibatch size of our ResNet50 based network is set to 8 due to the GPU memory limitation. The other hyperparameters are set as the same as the ones used in the VGG-16 based network. The testing time for one image is 0.236s.

\subsection{Using the CRF Post-processing}

When we adopt the CRF post-processing method, we use the same parameters and the same code used by \cite{hou2017dss}. It additionally costs another 0.09s for each image.

\section{Experiments}

\subsection{Effectiveness of Progressively Embedding PiCANets}

\begin{table} [!t]
\begin{center}
\caption{Effectiveness of progressively embedding PiCANets. ``+75G432LP'' means using \textbf{G}lobal PiCANets in $\mathcal D^7$ and $\mathcal D^5$, and \textbf{L}ocal \textbf{P}iCANets in $\mathcal D^4$, $\mathcal D^3$, $\mathcal D^2$. Other settings can be inferred similarly. \blu{Blue} indicates the best performance.}
\vspace{1mm}
\label{progressiveTab}
\footnotesize
\begin{tabular}{@{}lccccccc@{}}
\toprule
\multirow{2}{*}{Settings} & \multicolumn{3}{c}{DUT-O \cite{yang2013gbmr}}  && \multicolumn{3}{c}{DUTS-TE \cite{wang2017duts}} \\
\cmidrule{2-4} \cmidrule{6-8}
                          &  $F_{\beta}$   &  $F_{\beta}^{\omega}$  &   MAE    && $F_{\beta}$   &  $F_{\beta}^{\omega}$  &   MAE  \\ \midrule
U-Net \cite{ronneberger2015unet} & 0.761   & 0.651                  & 0.073    && 0.819         & 0.715                  & 0.060  \\ \midrule
+7GP                      & 0.772          & 0.660                  & 0.071    && 0.826         & 0.722                  & 0.058  \\
+75GP                     & 0.778          & 0.662                  & 0.071    && 0.834         & 0.724                  & 0.057  \\
+75G4LP                   & 0.785          & 0.678                  & 0.069    && 0.840         & 0.736                  & 0.056  \\
+75G43LP                  & 0.791          & 0.682                  &\blu{0.068}&& 0.848        & 0.740                  & 0.055  \\
+75G432LP                 &\blu{0.794}     &\blu{0.691}             &\blu{0.068}&&\blu{0.851}   &\blu{0.748}             &\blu{0.054}\\
\bottomrule
\end{tabular}
\vspace{-0.7cm}
\end{center}{}
\end{table}

Here we report a more detailed ablation study of progressively embedding PiCANets in each decoding module. As shown in Table~\ref{progressiveTab}, progressively embedding global and local PiCANets in $\mathcal D^7,\mathcal D^5,\mathcal D^4,\cdots,\mathcal D^1$ can consistently improve the saliency detection performance, thus demonstrating the effectiveness of our proposed PiCANets and the saliency detection model.

\subsection{More Visualization of the Learned Attention Maps}

\begin{figure*}[!t]
  \graphicspath{{Figures/attention/}}
  \centering
  \begin{overpic}[width=1\linewidth]{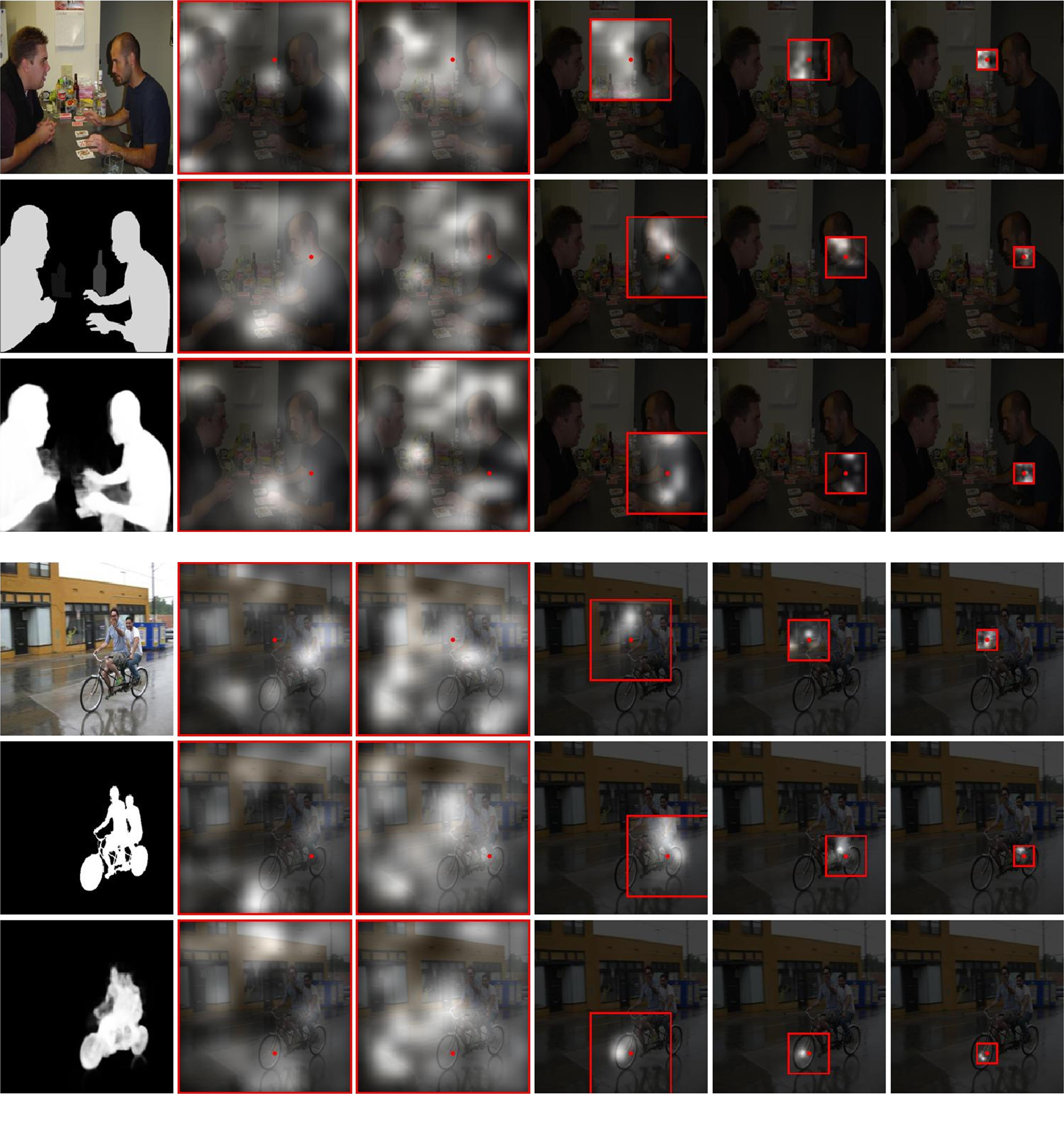}
  \put(2,1){\small Image/GT/SM}
  \put(21,1){\small att($\mathcal D^7$)}
  \put(37,1){\small att($\mathcal D^5$)}
  \put(53,1){\small att($\mathcal D^4$)}
  \put(69,1){\small att($\mathcal D^3$)}
  \put(85,1){\small att($\mathcal D^2$)}
  \end{overpic}
  \caption{Illustration of the learned attention maps of the proposed PiCANets. The first column shows two images and their corresponding ground truth masks and the predicted saliency maps of our model while the last five columns show the attention maps in five attended decoding modules, respectively. For each image, we give three example pixels (denoted as red dots. The first row shows a background pixel and the bottom two rows show two foreground pixels). The attended context regions are marked by red rectangles.}
  \label{attention_sup}
  \vspace{-0.4cm}
\end{figure*}

We illustrate more learned attention maps in Figure~\ref{attention_sup} for the five attended decoding modules. Figure~\ref{attention_sup} shows that the global attention learned in $\mathcal D^7$ and $\mathcal D^5$ can attend to foreground objects for background pixels and backgrounds for foreground pixels. The local attention learned in $\mathcal D^4$, $\mathcal D^3$, and $\mathcal D^2$ can attend to regions with similar semantics with the referred pixel.

\subsection{More Visual Comparison Between Our Model and Stata-of-the-Art Methods}

\begin{figure*}[!t]
  \graphicspath{{Figures/qualitative/}}
  \centering
  \begin{overpic}[width=1\linewidth]{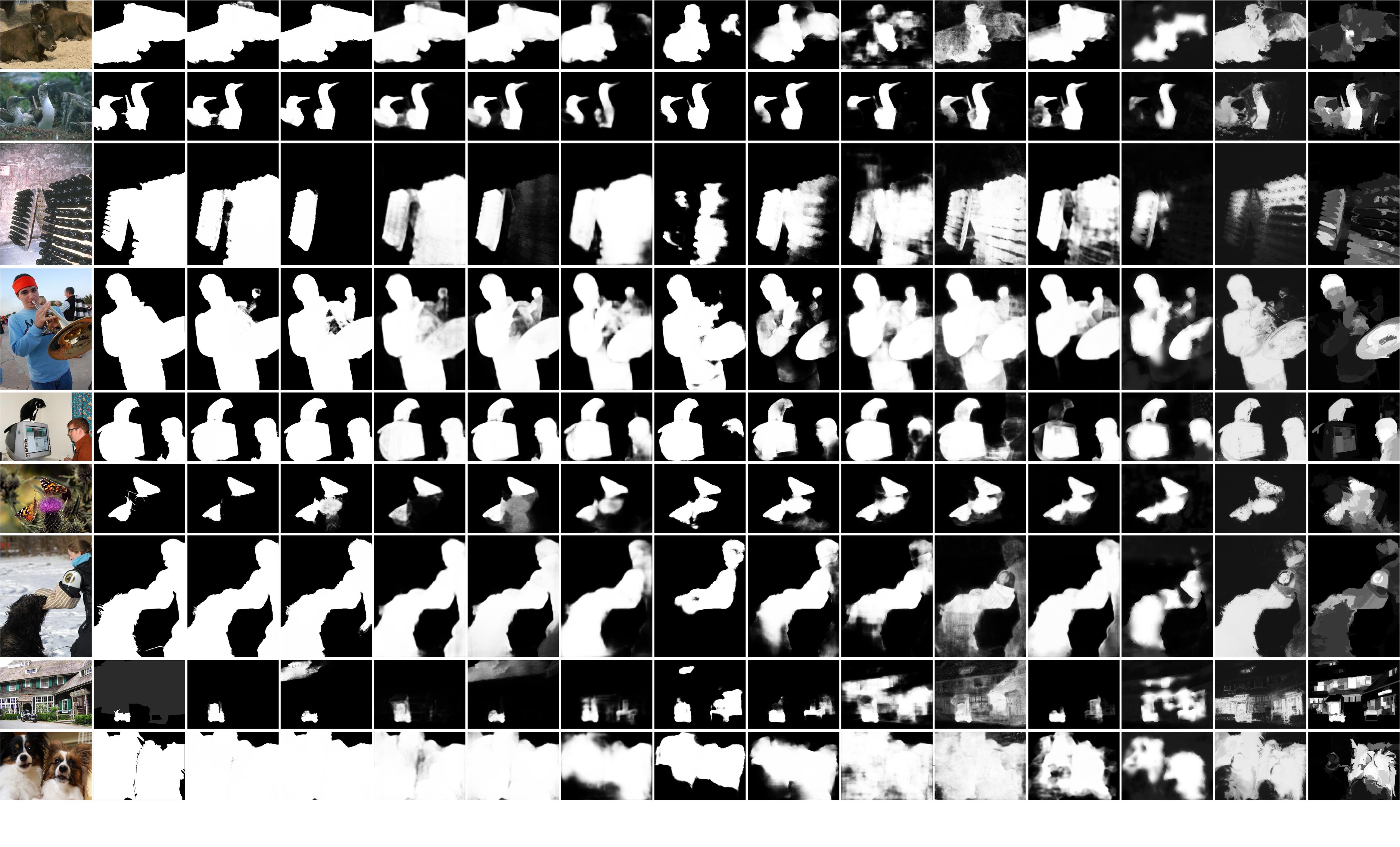}
  \put(1,1.5){\scriptsize Image}
  \put(9,1.5){\scriptsize GT}
  \put(13.8,0.5){\scriptsize \shortstack[c] {PiCANet\\ -RC}}
  \put(20.8,0.5){\scriptsize \shortstack[c] {PiCANet\\ -C}}
  \put(27.5,0.5){\scriptsize \shortstack[c] {PiCANet\\ -R}}
  \put(34.5,1.5){\scriptsize PiCANet}
  \put(40.5,1.5){\scriptsize SRM \cite{Wang2017srm}}
  \put(48,1.5){\scriptsize DSS \cite{hou2017dss}}
  \put(53.5,1.5){\scriptsize NLDF \cite{luo2017nldf}}
  \put(60,1.5){\scriptsize Amulet \cite{Zhang2017amulet}}
  \put(67.3,1.5){\scriptsize UCF \cite{Zhang2017ucf}}
  \put(74,1.5){\scriptsize DHS \cite{liu2016dhsnet}}
  \put(80.2,1.5){\scriptsize RFCN \cite{wang2016rfcn}}
  \put(87.6,1.5){\scriptsize DCL \cite{li2016dcl}}
  \put(94,1.5){\scriptsize MDF \cite{li2015mdf}}
  \end{overpic}
  \caption{Qualitative comparison. (GT: ground truth)}
  \label{visualcmp_sup}
\end{figure*}

We also show more qualitative results in Figure~\ref{visualcmp_sup}. It shows that compared with other state-of-the-art methods, our model can highlight salient objects more accurately and uniformly under various challenging scenarios even without using post-processing techniques.

\subsection{Failure Cases}

\begin{figure*}[!t]
  \graphicspath{{Figures/failureCases/}}
  \centering
  \includegraphics[width=1\linewidth]{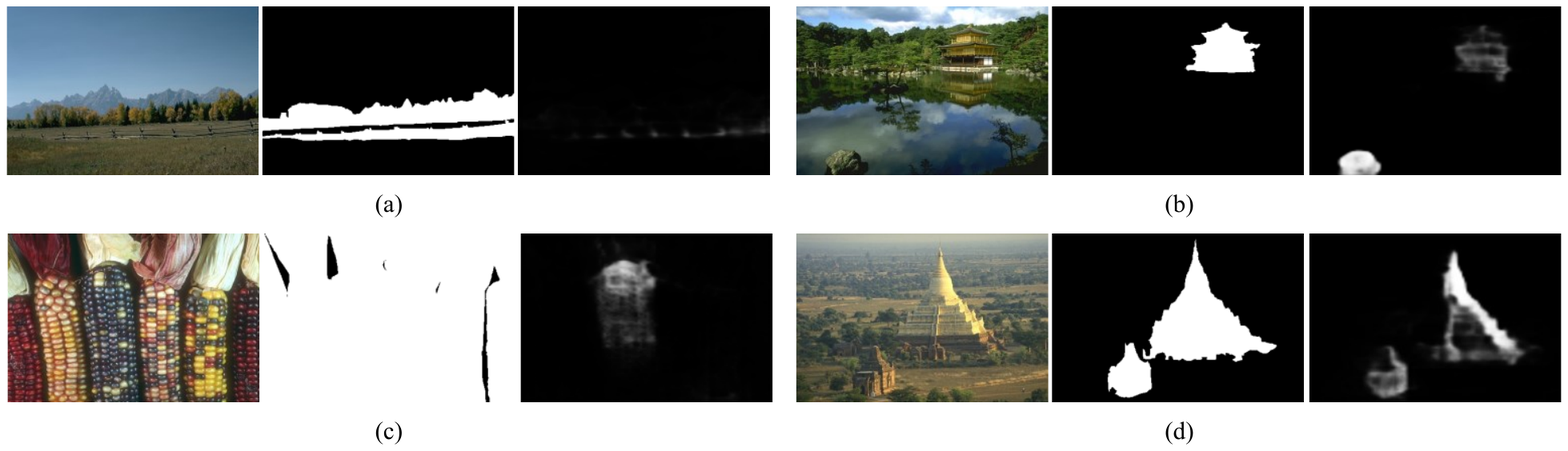}
  \caption{Failure cases. The three images in each set are the input image, the ground truth, and our result, respectively.}
  \label{failure_cases}
  \vspace{-0.3cm}
\end{figure*}

We show some failure cases of our PiCANet-R model in Figure~\ref{failure_cases}. Basically, our model usually fails when the image has no obvious foreground objects, as shown in (a) and (b). (c) shows that when the foreground object is extremely large, our model is also easy to fail. While these two situations are also challenging to other traditional and deep learning based saliency models, indicating that we still have much room to improve current models. (d) shows that the non-uniform illumination on the object may also mislead our model.

{\small
\bibliographystyle{ieee}
\bibliography{cvpr18_PiCANet_saliency}
}